# FinVector-Market-4B: A Controlled Study of LoRA Adaptation for Structured Financial Tasks

Alina Khaybullina

Independent Researcher
ORCID 0009-0007-2586-842X

October 2026

**Abstract**

FinVector-Market-4B adapts Qwen/Qwen3.5-4B with rank-16 LoRA on a 22,000-example corpus for structured financial tasks. We evaluate the base and adapted models on the same 600-example benchmark under implicit and explicit JSON-schema contracts. Supplying the schema alone raises base-model JSON validity from 0% to 91.3%. Under matched explicit prompting, the frozen scores improve from 14.7% to 40.0% for FinQA answer exact match, from 48.0% to 82.7% for calculator-expression correctness, from 20.1% to 89.5% for scenario branch-label agreement, and from 52.4% to 87.2% for implication-direction agreement. A post-hoc policy-scoring audit shows that the reported macro-F1 decline reflects a changing label set; using the same three target classes gives 77.4% for the base and 83.1% for the adapter. Filing overlap and calculator-target inconsistencies qualify the benchmark's generalization claims. The results show that compact financial domain adaptation can produce substantial task-specific gains beyond output-format learning under matched prompting, with gains bounded by the evaluated task distribution and prompt contract.

## 1 Introduction

Compact open-weight language models are increasingly attractive for domain-specific applications because they can be adapted efficiently, deployed with greater control, and evaluated under fixed task contracts. Finance is a demanding setting for this approach: useful systems must combine technical language, numerical reasoning, policy interpretation, tool use, and cross-asset causal structure while producing outputs that are stable enough to integrate with downstream software. FinVector-Market-4B was built to test how far a 4B-class model can be specialized for that combination of requirements.

The goal is not to build a general finance chatbot or a price-prediction model. FinVector targets eight auditable task families: policy tone, financial calculation, tool selection, tool calls, conditional reasoning, transmission chains, scenario trees, and structured market implications. Exact arithmetic is treated as a deterministic tool problem, while current market facts would belong in retrieval or external data services. The model is trained primarily for interpretation, decomposition, routing, and structured response generation.

Prior work has shown that finance-specific training can improve language-model performance, from large-scale domain pretraining such as BloombergGPT [1] to data-centric adaptation frameworks such as FinGPT [2]. FinQA established financial numerical reasoning over heterogeneous reports as a distinct challenge [3], while LoRA made parameter-efficient specialization practical for large pretrained models [4]. At the same time, prompt-sensitivity and benchmark research shows that instruction phrasing, output format, and scoring contracts can materially alter measured performance [5,6,8-10]. A credible fine-tuning study therefore has to separate what the model learned about the task from what it learned about how answers must be serialized.

FinVector-Market-4B adapts Qwen/Qwen3.5-4B [7] at the pinned revision listed in Table 8 using rank-16 BF16 LoRA. The 22,000-example corpus combines central-bank communications, FinQA, deterministic FinMath problems, and structural macro-to-market tasks. Source records are transformed into conversational supervision; provenance, licensing metadata, task family, and split-group identifiers remain outside the model input.

The evaluation crosses model state with prompt contract in four frozen arms. Giving both models the same explicit task schema reduces a major interface confound, while the implicit arms retain the combined effects of learned task behavior, response format, and termination. The control does not fully separate semantic reasoning from lexical target conventions or familiarity with the generator distribution.

We address three research questions:

**Domain adaptation:** Can a compact 4B-class model be specialized for structured financial reasoning with a small LoRA adapter while keeping the base model frozen?

**Prompt compliance:** How much of the apparent improvement over the base model is explained by learning the required output contract rather than by task-specific adaptation?

**Residual task gains and prompt sensitivity:** Which improvements remain under a matched explicit-schema control, and which metrics or task families regress or remain sensitive to prompt formulation?

The paper contributes a documented financial supervision pipeline, a parameter-efficient Qwen3.5-4B adaptation with 21.23 million trainable parameters, and a four-arm comparison under fixed decoding and scoring conditions. It also reports post-hoc metric and dataset audits that qualify the interpretation of the frozen results. The grouping guarantees concern the identifiers actually used by the pipeline; they do not establish filing-level or template-level independence.

Explicit schemas substantially improve the base model's structured-output scores, and additional gains remain after adaptation. The adapted model remains prompt-sensitive. The policy macro-F1 result also illustrates why scoring definitions require scrutiny: an invalid-output label changes the registered metric's averaging denominator, creating an apparent decline that disappears when the same target classes are used in both arms.

## 2 Related Work

### 2.1 Financial language models

BloombergGPT [1] demonstrated the value of domain-specific language-model training using a large mixture of financial and general-domain text, reporting gains across public financial tasks and internal benchmarks. FinGPT [2] pursued an open-source, data-centric alternative, emphasizing accessible financial data pipelines and lightweight adaptation techniques. FinVector differs in objective and scale: it does not attempt broad financial pretraining. Instead, it studies whether a compact 4B-class model can be adapted for a narrow set of auditable reasoning and structured-output tasks, and whether measured gains persist under a prompt-format control.

### 2.2 Financial numerical reasoning

FinQA [3] provides financial-report evidence, expert-authored questions, and executable reasoning programs. FinVector transforms a sample from the upstream training partition and applies its own group split; its 75-question benchmark slice is not the official FinQA leaderboard test. We report normalized answer exact match rather than the official execution-accuracy or program-accuracy measures, so direct leaderboard comparisons would be inappropriate.

### 2.3 Parameter-efficient adaptation

LoRA freezes pretrained weights and learns low-rank updates [4]. FinVector uses rank 16, alpha 16, and dropout 0.0. The released adapter contains 21,233,664 parameters across 128 module pairs: feed-forward projections in all 32 decoder layers and query, key, value, and output projections in the eight full-attention layers. The underlying Qwen3.5 architecture also contains linear-attention layers, whose attention modules are not adapted by these keys. Training used BF16 on an H100; the converted MLX release stores its adapter tensors in FP32.

### 2.4 Prompt sensitivity and evaluation confounds

Instruction-tuned models remain sensitive to the phrasing of otherwise appropriate instructions [6], and prompt-formatting choices can produce large score differences even when semantic content is preserved [5]. For that reason, the paper uses an explicit-schema control. Rather than interpreting the base model's failure to satisfy the expected structured contract as zero latent reasoning capability, we evaluate a second base arm in which the task-specific JSON shape is supplied directly. The same pair of prompt contracts is then applied to the fine-tuned adapter.

### 2.5 Benchmark validity and behavioral evaluation

Benchmark design can alter what a model score appears to measure. Bowman and Dahl [8] argue that language-understanding benchmarks require clearer alignment between benchmark construction and the capabilities they are intended to measure. Raji et al. [9] similarly emphasize construct-validity problems when benchmark performance is treated as a broad proxy for general capability. Ribeiro et al. [10] show that behaviorally targeted tests can expose failure modes hidden by aggregate held-out accuracy. These concerns are directly relevant to structured-output fine-tuning, where a score may reflect both substantive task behavior and familiarity with an answer contract.

The present study combines these lines of work in a domain-adaptation setting. Finance-specific language models test whether specialization changes task performance; FinQA provides grounded numerical reasoning; LoRA provides an efficient adaptation mechanism; and prompt-sensitivity research shows that semantically equivalent interfaces can shift measured scores. FinVector adds a controlled comparison in which the benchmark is held fixed while model state and prompt-contract visibility are varied independently.

## 3 FinVector Corpus

The frozen V1 corpus contains 22,000 examples covering the target FinVector behaviors. Externally grounded records are combined with deterministic financial calculations and synthetic structural market tasks. These components provide task-specific supervision rather than a broad financial pretraining distribution.

### 3.1 Composition

**Table 1.** Corpus composition and recorded license provenance.

| Component | Rows | Share | Grounding | Recorded licence |
|---|---|---|---|---|
| Central-bank comms. | 4,500 | 20.45% | External | CC-BY-4.0 |
| FinQA | 3,500 | 15.91% | External | CC-BY-4.0 mirror card; MIT upstream repository |
| FinVector FinMath | 6,500 | 29.55% | Synthetic / project-generated | Project-generated |
| FinVector Implications | 7,500 | 34.09% | Synthetic / project-generated | Project-generated |
| Total | 22,000 | 100.00% | 8,000 external / 14,000 synthetic | Mixed |

Externally grounded records account for 36.36% of the corpus. The central-bank source is pinned to the aufklarer/central-bank-communications revision recorded in [11]; FinQA is obtained from the upstream commit recorded in [13]. IBM's mirror card lists CC-BY-4.0 [12], while the pinned upstream repository carries an MIT license [13]; these are recorded as separate provenance facts.

Central-bank annotations are mapped into classification conversations: tightening, hike, and hawkish labels map to hawkish; easing, cut, and dovish labels map to dovish; rate-hold and neutral labels map to neutral. FinQA conversations include source report text, tables, and questions, with answers and programs as targets. FinMath and structural examples are generated programmatically. All records use a common system/user/assistant format, with audit metadata excluded from the model input.

### 3.2 Task families

The corpus maps into eight registered task families:

**Table 2.** Task-family counts across all corpus splits.

| Task family | Rows |
|---|---|
| Policy tone | 4,500 |
| Financial calculation | 3,500 |
| Tool selection | 3,250 |
| Tool call | 3,250 |
| Conditional reasoning | 1,875 |
| Transmission chain | 1,875 |
| Scenario tree | 1,875 |
| Probabilistic market implications | 1,875 |

The structural tasks follow a common reasoning path: information to macro shock to policy effect to transmission channel to cross-asset implication to alternative scenario or uncertainty. The generated examples are used as task-specific supervision rather than as evidence about realized market returns.

### 3.3 Splits and quality gates

The split contains 17,489 training, 2,211 validation, and 2,300 test rows. A 600-row benchmark is selected deterministically from the test partition, with 75 examples per task. A hash-based assignment of group identifiers, using seed 3407 and nominal 80/10/10 proportions, keeps each defined group in one split. Central-bank groups use document identifiers; FinMath groups join views of the same problem; structural groups join views of the same event. FinQA identifiers retain the source page path after removing the question suffix, so the implemented grouping is page-level rather than annual-filing-level.

22,000 unique row IDs out of 22,000;
22,000 unique normalized fingerprints out of 22,000;
zero duplicate IDs and zero duplicate normalized fingerprints;
zero overlap of registered group identifiers across train, validation, and test;
zero invalid canonical training records and zero task-schema-invalid target objects;
zero probability-invalid rows.

These checks were reproduced from the locked files. They establish integrity under the implemented grouping, not complete protection against shared source context. The filing-level audit in Section 7.3 identifies FinQA overlap across splits. The benchmark remains unchanged; its SHA-256 is listed in Table 8.

## 4 Experimental Design

### 4.1 Four frozen arms

The evaluation crosses model state and prompt contract:

**Table 3.** Four frozen evaluation arms.

| | Implicit FinVector contract | Explicit task JSON schema |
|---|---|---|
| Base Qwen3.5-4B | FV-BASE-001 | FV-BASE-002 |
| FinVector adapter | FV-SFT-001-EVAL | FV-SFT-001-EVAL-EXPLICIT |

All four arms use the same 600 examples, base-model revision, seed 3407, greedy decoding, disabled thinking, batch size 8, and a 700-new-token ceiling. Evaluation runs on Apple Metal with MLX 0.32.2 and MLX-LM 0.31.3. The runner trims surrounding whitespace, then scores the generated text without Markdown removal, answer extraction, JSON repair, or key rewriting.

FV-BASE-001 asks the base model to infer the expected structured answer contract. Its score combines task behavior, output-format inference, verbosity, and termination behavior. FV-BASE-002 changes only the prompt contract by appending the task-specific JSON schema. Both model states use the same frozen schema registry, identified in Table 8.

The post-training arms use the same two contracts. This produces the central comparison of the paper:

BASE-002 explicit vs. SFT explicit asks whether gains remain after both models are told the output schema.

The explicit comparison reduces the confound of unequal schema visibility. It does not isolate every consequence of supervised fine-tuning: score differences can still reflect learned wording, branch conventions, response length, generator patterns, or semantic task behavior. We therefore interpret residual differences as performance on matched task contracts rather than a complete decomposition of reasoning and formatting.

The benchmark and base controls were frozen before the completed training run, and the reported results were not used to retune the final model or benchmark. Internal records designate the implicit contract as the original primary comparison. This article emphasizes the explicit comparison for interpretation while retaining both contracts and distinguishing that analytic emphasis from a new preregistration. The scoring and dataset audits in Section 7 are post-hoc analyses of unchanged artifacts.

The explicit-schema arm tests whether the observed gap disappears when the base model is told the JSON shape. Its perfect tool-selection score is a useful ceiling control. Remaining score differences support specialization to the evaluated contracts, but the dataset and scoring limitations described below prevent interpreting all of those differences as general financial reasoning gains.

### 4.2 Metrics

The frozen scorer reports raw JSON parsing validity, task-schema validity, exact target-object match, policy-tone accuracy and macro-F1, FinQA answer exact match, calculator metrics, exact causal-component overlap, scenario branch-label agreement, and implication-direction agreement. Exact JSON match compares parsed objects, so object-key order is irrelevant. FinQA answers are compared after whitespace normalization and case-folding; numerical program execution is not part of that metric.

Calculator-expression correctness means that the generated expression executes and returns a value within an absolute tolerance of 0.011 of the stored answer. The separately scored answer field uses the same tolerance. Neither measure requires textual expression identity. Policy macro-F1 averages class F1 scores over the union of observed target and prediction labels; parse failures therefore introduce an additional class. Section 7.2 retains that frozen rule and adds a fixed-class diagnostic.

Scenario scores pool branch-asset labels over 750 components: 375 directional labels and 375 labels marked conditional. Missing components or unmatched branch conditions fail. Implication scores pool the largest-probability direction over 525 asset components. These components are clustered within 75 examples per task and are not independent observations.

Probability calibration is not measured, because the benchmark contains structural probability priors rather than realized-outcome labels.

Causal-component F1 pools exact normalized list-item overlaps within each task. It is a lexical micro-F1 measure, not a semantic or human assessment of causal validity.

## 5 Model Adaptation Procedure

FV-SFT-001 uses BF16 LoRA on the pinned Qwen/Qwen3.5-4B revision. Training uses Unsloth and PEFT; the resulting adapter is converted to the MLX-LM layout for evaluation and release.

### 5.1 Configuration

**Table 4.** Final training configuration.

| Parameter | Value |
|---|---|
| Method | BF16 LoRA |
| LoRA rank / alpha / dropout | 16 / 16 / 0.0 |
| Target modules | MLP: all 32 layers;<br>q, k, v, o: 8 full-attention layers |
| Trainable LoRA parameters | 21,233,664 |
| Max sequence length | 4,096 |
| Padding | Dynamic, longest in batch |
| Assistant-only loss | Yes |
| Packing | No |
| Epochs | 1 |
| Batch size | 8 |
| Gradient accumulation | 1 |
| Effective batch | 8 examples |
| Optimizer | AdamW 8-bit |
| Maximum learning rate | 1e-4 |
| Warmup | 3% |
| Observed schedule | Linear warmup, then linear decay |
| Seed | 3407 |

The training split contains 6,737,177 unpadded tokens. The exact-tokenizer audit finds no training or validation conversation longer than the 4,096-token limit. Evaluation is configured once at epoch end, so only one validation-loss observation exists.

### 5.2 Training runtime and hardware

The completed run used one NVIDIA H100 80GB HBM3 and executed 2,187 optimizer steps. Trainer runtime was 1,904.73 seconds (31.75 minutes); the provider job took 1,992.94 seconds (33.22 minutes), including additional work around training.

### 5.3 Training trajectory

The aggregate training loss is 0.183375 and the single epoch-end validation loss is 0.138475. Figure 1 plots the 437 recorded training-loss events and a trailing 15-record mean.

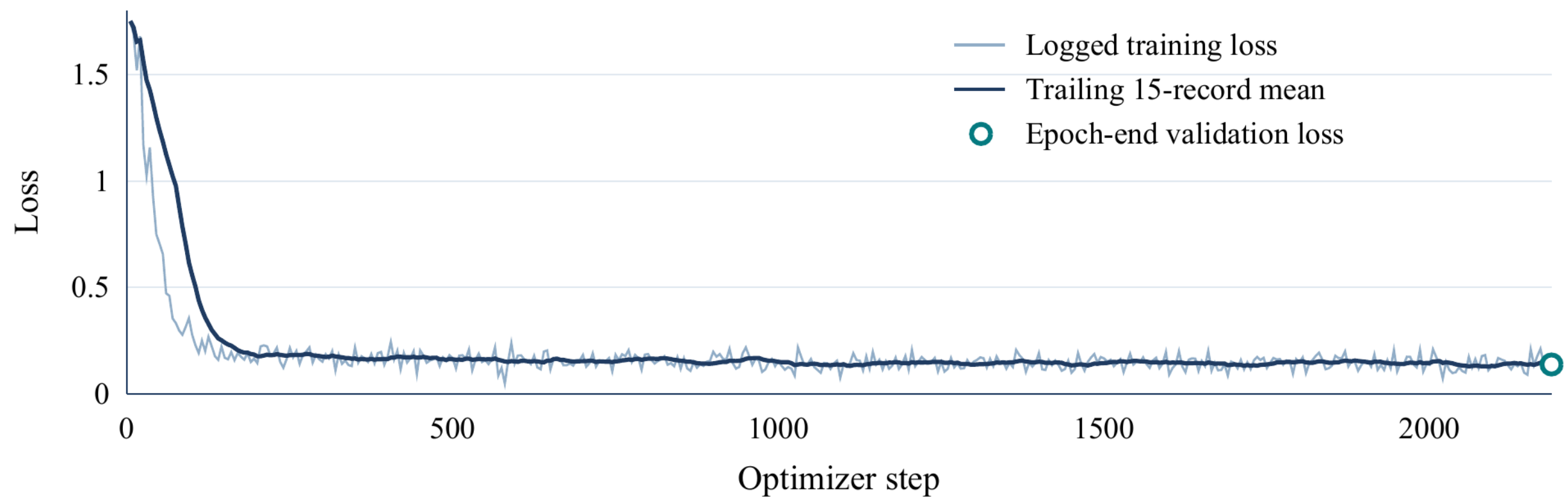

**Figure 1.** Recorded FV-SFT-001 training losses, a trailing 15-record mean, and the single epoch-end validation observation. The mean uses available records at the start of the trace.

# 6 Results

## 6.1 Why the explicit-schema comparison is emphasized

Under the implicit contract, the base model fails the strict structured-output criterion: JSON validity is 0%, and 280/600 outputs reach the 700-token ceiling. Supplying the schema alone changes the measured baseline substantially: FV-BASE-002 reaches 91.33% JSON validity and 90.33% task-schema validity, with 23/600 ceiling hits. This demonstrates that the original zero baseline is dominated by interface mismatch and verbosity as well as task performance.

The explicit comparison is emphasized because it reduces the schema-visibility confound. The implicit pair remains the originally designated primary comparison in the internal protocol.

## 6.2 Matched explicit-schema comparison

**Table 5.** Matched explicit-schema results, with a labeled post-hoc policy diagnostic.

| Metric | Base explicit | FinVector explicit | Change |
|---|---|---|---|
| Strict JSON validity | 91.33% | 99.83% | +8.50 pp |
| Full task-schema validity | 90.33% | 99.83% | +9.50 pp |
| Exact JSON match | 0.00% | 43.00% | +43.00 pp |
| Policy-tone accuracy | 80.00% | 84.00% | +4.00 pp |
| Registered policy macro-F1* | 77.35% | 62.30% | -15.05 pp |
| Policy macro-F1, fixed classes† | 77.35% | 83.07% | +5.71 pp |
| FinQA answer exact match | 14.67% | 40.00% | +25.33 pp |
| Tool-selection accuracy | 100.00% | 100.00% | 0.00 pp |
| Executable calculator expression | 84.00% | 100.00% | +16.00 pp |
| Correct calculator expression | 48.00% | 82.67% | +34.67 pp |
| Calculator answer tolerance | 8.00% | 44.00% | +36.00 pp |
| Conditional causal-component F1 | 0.00% | 99.95% | +99.95 pp |
| Transmission-chain component F1 | 0.00% | 100.00% | +100.00 pp |
| Scenario branch-label agreement | 20.13% | 89.47% | +69.33 pp |
| Implication schema validity | 92.00% | 100.00% | +8.00 pp |
| Implication-direction agreement | 52.38% | 87.24% | +34.86 pp |

* The registered scorer averages over observed labels, including parse failures. † Post-hoc recomputation over the same three policy classes; malformed predictions still count as errors. See Section 7.2. Changes use unrounded values.

FinQA and calculator metrics improve under the frozen scorer, while explicit tool-selection accuracy remains at its 100% ceiling. The registered policy macro-F1 comparison uses different observed label sets and needs the qualification in Section 7.2. Filing overlap, calculator-target precision, and the conditional-label component of the scenario score further bound the interpretation of these gains (Section 7.3).

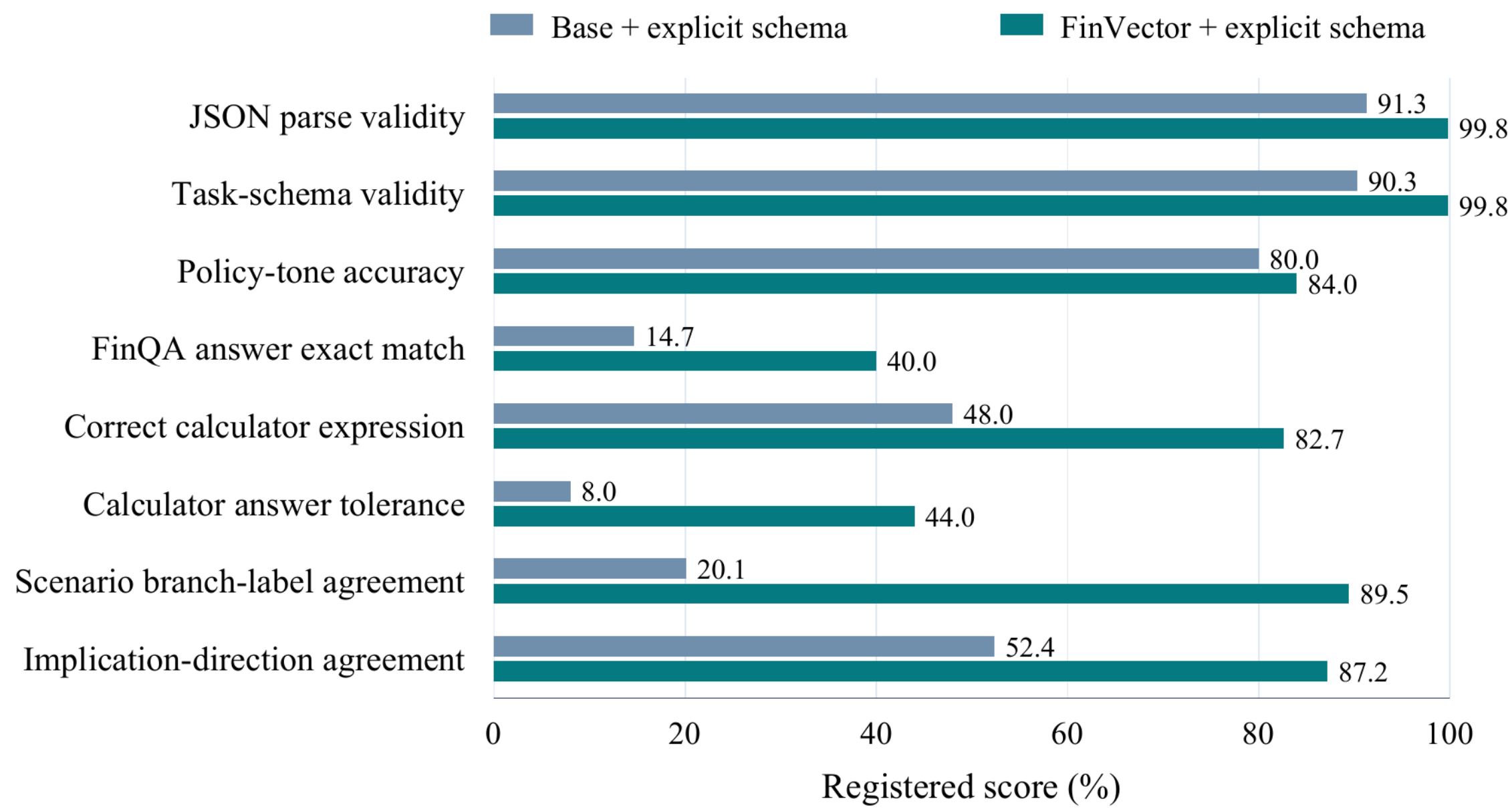

**Figure 2.** Selected frozen scores under matched explicit-schema prompting. FinQA uses the project's own held-out slice; calculator scores use stored targets; scenario scores include conditional branch labels. Section 7.3 reports supplementary validity checks. Macro-F1 is omitted here because its registered label set varies across arms.

### 6.3 Implicit contract

Under the implicit contract, FinVector produces valid JSON on all 600 examples and no output reaches the 700-token ceiling. Exact JSON match is 47.33%. Policy-tone accuracy reaches 86.67%, FinQA exact match 40.00%, calculator-expression correctness 85.33%, scenario branch-label agreement 97.73%, and implication-direction agreement 96.38%.

The implicit results describe performance under the learned FinVector contract. They combine task behavior with answer-format and termination behavior, and are not an isolated estimate of domain specialization.

### 6.4 Output length and contract learning

Output length provides a further contract diagnostic. The number of generations reaching the 700-token ceiling is:

BASE-001 implicit: 280/600
SFT implicit: 0/600
BASE-002 explicit: 23/600
SFT explicit: 1/600

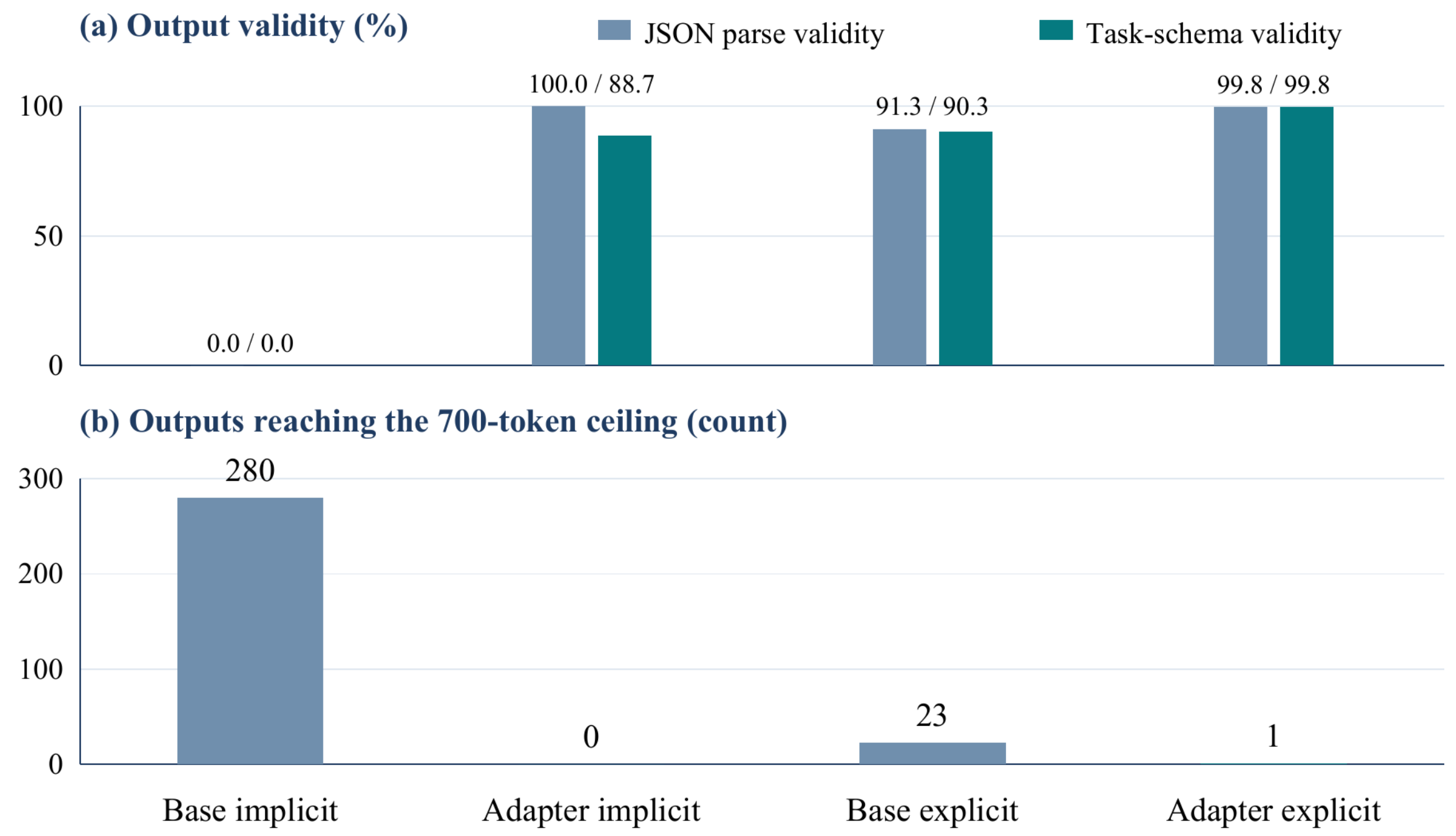

**Figure 3.** Four-arm format and termination diagnostics (600 outputs per arm). Panel (a) shows JSON parse and task-schema validity; labels give the two percentages in that order. Panel (b) counts outputs reaching the 700-token ceiling. Reaching the ceiling is a length indicator; it does not by itself prove that every such response lacks a usable answer.

Explicit schemas reduce ceiling hits from 280 to 23 for the base model. The adapter reaches the ceiling on zero implicit and one explicit output. These counts document a substantial change in response-length behavior under the benchmark contracts.

## 7 Error Analysis

### 7.1 Implicit tool-selection failures are mostly contract omissions

The implicit adapted arm records 45/75 tool-selection matches. The other 30 outputs are valid JSON and select calculator, but omit the required boolean tool_required. The registered metric requires both fields, so those rows fail. Under the explicit schema, the adapted model reaches 75/75. Every tool-selection target requires a calculator; the task therefore measures this contract's completion rather than discrimination between tool and no-tool cases.

This case distinguishes contract completion from simple selection of the uniformly positive calculator target. It also illustrates that the adapter remains prompt-sensitive despite training on structured outputs.

### 7.2 Policy macro F1 and the averaging label set

Policy accuracy rises from 60/75 to 63/75, while registered macro-F1 falls from 0.773521 to 0.622989. The frozen scorer uses every label found in either targets or predictions. The baseline has three such labels; the explicit adapted arm has an additional __parse_failure__ label with zero F1. Consequently, the two values average over different label sets.

**Table 6.** Explicit policy-tone confusion counts. The malformed response remains an error.

| Target class | Base explicit predictions | FinVector explicit predictions |
|---|---|---|
| Dovish (26) | 19 dovish, 3 hawkish, 4 neutral | 26 dovish |
| Hawkish (18) | 10 hawkish, 1 dovish, 7 neutral | 11 hawkish, 4 dovish, 3 neutral |
| Neutral (31) | 31 neutral | 26 neutral, 4 dovish, 1 malformed/truncated |

Recomputing macro-F1 over the same three target classes gives 0.773521 for the base and 0.830651 for the adapter. The malformed neutral prediction remains a false negative for neutral; it is not removed. Per-class F1 increases from 0.8261 to 0.8667 for dovish, from 0.6452 to 0.7586 for hawkish, and from 0.8493 to 0.8667 for neutral. The published 0.622989 equals three-fourths of the adapted model's fixed-three-class score.

The registered score is reproduced correctly under the frozen implementation, but its decline does not establish deteriorating class-balanced policy classification. The fixed-class calculation is a post-hoc diagnostic, reported alongside the original result rather than replacing it. Prompt sensitivity remains: the implicit adapted arm's fixed-class macro-F1 is 0.859333, above the explicit arm's 0.830651.

### 7.3 Post hoc dataset and component audits

The retained dataset and all four raw-output files were audited without altering their bytes or rerunning generation. FinQA has no page-group overlap, but 112 annual filings occur in both training and test. Of the 75 benchmark questions, 53 share a filing with training data. On the 22 questions from unshared filings, exact match is 2/22 for the explicit base and 8/22 for the explicit adapter. This smaller, post-hoc subset is informative but does not replace a prospectively filing-disjoint evaluation.

Calculator targets use high-precision sampled inputs while questions display rounded values. Across the 3,250 tool-call records, 643 stored answers differ from calculations using the displayed inputs by more than 0.011; 11 such records occur in the 75-item benchmark. Removing those 11 items as a post-hoc diagnostic leaves 64 examples. Expression correctness is 36/64 for the explicit base and 62/64 for the explicit adapter; answer-field correctness is 5/64 and 33/64. These subset results retain the original targets and tolerance and do not redefine the registered scores.

The scenario score also combines distinct behaviors. Half of its 750 components are fixed conditional labels. On the 375 up, neutral, or down components alone, the explicit base matches 151/375 and the explicit adapter 296/375; the remaining 375 conditional components score 0/375 and 375/375. The aggregate increase therefore includes learning the benchmark branch convention. The implicit adapter matches 358/375 directional components. All of these are synthetic-label comparisons, with components nested within examples.

**Table 7.** Supplementary post-hoc diagnostics under explicit prompting. These analyses do not replace frozen endpoints.

| Diagnostic | Base | FinVector |
|---|---|---|
| Policy macro-F1, same three classes | 77.35% | 83.07% |
| FinQA, unshared filings (22 questions) | 2/22 (9.09%) | 8/22 (36.36%) |
| Expression correctness, 64 consistent targets | 36/64 (56.25%) | 62/64 (96.88%) |
| Answer tolerance, 64 consistent targets | 5/64 (7.81%) | 33/64 (51.56%) |
| Scenario directional components (375) | 151/375 (40.27%) | 296/375 (78.93%) |
| Scenario conditional components (375) | 0/375 (0.00%) | 375/375 (100.00%) |

## 8 Discussion

### 8.1 What the prompt control reveals

The implicit zero baseline combines interface mismatch and verbosity with task behavior. Supplying the schema to the unchanged model raises JSON validity to 91.3%, policy accuracy to 80%, and tool-selection accuracy to 100%. This shows that measured zero-shot failure can substantially understate performance under a specified interface.

A comparison using only the implicit base would overstate what can be attributed to domain adaptation. The explicit control improves the interpretation of the experiment, while lexical targets and generator-specific contracts remain additional sources of score variation.

### 8.2 Fine-tuning gains remain after schema knowledge is supplied

Residual frozen-score gains include 25.33 percentage points in FinQA exact match, 34.67 points in calculator-expression correctness, 36.00 points in calculator answer tolerance, 69.33 points in scenario branch-label agreement, and 34.86 points in implication-direction agreement. Both explicit arms receive the same schema, but shared filing context and learned target conventions remain relevant explanations for part of the observed differences.

The strongest defensible interpretation is specialization to the evaluated task distribution. The supplementary subsets continue to favor the adapter, but they are post-hoc and smaller than the original benchmark. The experiment does not identify how much of each gain transfers to unseen filings, new templates, independently authored causal explanations, or realized market outcomes.

### 8.3 Parameter-efficient specialization and task trade-offs

Updating 21.23 million LoRA parameters changes behavior substantially while leaving the base weights frozen. Gains remain dependent on the prompt and scoring contract. For policy tone, comparable fixed-class F1 improves in the explicit pair, while the implicit adapted prompt retains the higher score. This illustrates both the effect of compact adaptation and the need to compare metrics with consistent definitions.

### 8.4 Structural market reasoning is not forecasting

FinVector's scenario and implication tasks encode conditional structural relationships. High direction-agreement scores show consistency with those registered targets; they do not show equivalent accuracy in predicting realized asset returns. Synthetic scenario probabilities are structural weights rather than calibrated forecast probabilities. A forecasting claim would require timestamped events, contemporaneous expectations, realized outcomes, and proper probabilistic scoring rules such as Brier score or log loss.

## 9 Limitations

This study has several important limitations.

**Benchmark scope.** The 600-example benchmark covers eight structured financial tasks. It should not be interpreted as a broad measure of language-model capability, financial expertise, or deployment suitability.

**Synthetic structural targets.** Scenario and market-implication scores measure agreement with synthetic labels rather than realized market returns. Probability calibration is not evaluated.

**Prompt sensitivity.** The adapted model's policy F1, tool selection, calculator answers, and structural scores vary with the prompt contract. No single score should be treated as prompt-independent model quality.

**Lexical causal metric.** Causal-component F1 uses exact normalized list-item overlap. Near-perfect scores do not establish semantic causal correctness without a separate semantic or human evaluation.

**Source concentration.** Central-bank examples focus on monetary policy, and FinQA has an imbalanced operation distribution. These characteristics limit the breadth of the conclusions.

## 10 Materials and Reproducibility

The public release is an MLX-LM LoRA adapter for the pinned Qwen/Qwen3.5-4B model, not a merged model or a directly loadable PEFT release. The public weight file matches the adapter hash recorded in the frozen MLX evaluations. Its SHA-256 is:

```
8defb30088c1a4a6220d7e6d6ab3df5305a419509b5c5e6d274fdbe0fe7e02a5
```

Public resources:

Hugging Face model: https://huggingface.co/alinakhay/FinVector-Market-4B

Hugging Face demo: https://huggingface.co/spaces/alinakhay/FinVector-Market-4B-Demo

GitHub project: https://github.com/alinakhay/FinVector-Market-4B

The corpus files, benchmark, evaluation outputs, and completed training artifacts were checked against their retained manifests during the post-hoc audit. Public resources [14] expose the adapter, usage example, and aggregate scores; the locked corpus, prompts, raw outputs, and proprietary construction code remain private. Public inference can be reproduced, but independent item-level benchmark reproduction requires those private materials.

### 10.1 Frozen identities

**Table 8.** Frozen artifact identifiers. Hashes identify the retained files, not public access to private data.

| Artifact | SHA-256 / identifier |
|---|---|
| Dataset manifest | d0f3cb99d699b23eb97f87de6bdd2da5f1e347330da519714ea31c7f11179f79 |
| Benchmark | 1a86f468d74dba7f6605d7d70a69cc61a2336bfc6603283f72ba3c108c752d04 |
| Train split | dfb4f2a6baa56df6896dc04fc82aacd6563f931923a9aac439f9e103853bc213 |
| Validation split | d1b0270c8799a8a4acd40d3e038e3b442141db6c061178969250c3d6ad9b4616 |
| Base model revision | 851bf6e806efd8d0a36b00ddf55e13ccb7b8cd0a |
| Schema registry | a31e3af9a1153a7e99bd6a0e89ea7bccca443f096ac2ff59259cf1b9087c454a |
| Training launch commit | a364de0d0bb9cf3fe4b975fce8d347db084e9cd3 |
| MLX adapter weights | 8defb30088c1a4a6220d7e6d6ab3df5305a419509b5c5e6d274fdbe0fe7e02a5 |

## 11 Conclusion

FinVector-Market-4B provides a compact adaptation and a controlled comparison for structured financial tasks. Explicit schemas recover substantial base-model performance, and the adapter scores higher on several metrics under the same explicit contract. The artifact audit reproduces the frozen results and verifies the retained dataset, evaluation, and training identities.

The validity audits qualify the inference from those scores. The apparent policy macro-F1 regression is explained by a changing averaging label set; a fixed-three-class diagnostic favors the adapter. FinQA filing overlap, hidden precision in calculator targets, and conditional components in the scenario metric limit broader generalization claims. Supplementary subsets remain favorable to the adapter but do not replace a newly registered evaluation on corrected data.

The evidence supports adaptation to the evaluated task contracts and synthetic target distribution. It does not establish universal financial expertise, independently validated causal reasoning, market prediction accuracy, or empirically calibrated probabilities.

Future work should register a new filing-disjoint and temporally appropriate benchmark, generate numerical targets from the exact displayed inputs, freeze the classification label set, separate conditional labels from directional outcomes, and evaluate independent templates and human-rated causal explanations. Additional seeds and models would test robustness; realized-outcome data would be required for a forecasting or calibration study.